\documentclass[10pt, a4paper]{article}
\usepackage{lrec2022} 
\usepackage{multibib}

\usepackage{graphicx, tabularx, soul, titlesec,epstopdf}

\titleformat{\section}{\normalfont\large\bfseries\center}{\thesection.}{1em}{}
\titleformat{\subsection}{\normalfont\SmallTitleFont\bfseries\raggedright}{\thesubsection.}{1em}{}
\titleformat{\subsubsection}{\normalfont\normalsize\bfseries\raggedright}{\thesubsubsection.}{1em}{}
\renewcommand\thesection{\arabic{section}}
\renewcommand\thesubsection{\thesection.\arabic{subsection}}
\renewcommand\thesubsubsection{\thesubsection.\arabic{subsubsection}}
\usepackage[utf8]{inputenc}

\usepackage{hyperref}
\usepackage{xstring}

\usepackage{color}
\usepackage{xspace}

\newcommand{\gm}{\textsc{Grew-match}\xspace}
\newcommand{\grew}{\textsc{Grew}\xspace}

\newcommand{\edge}[1]{\textbf{#1}}

\title{Graph Querying for Semantic Annotations}

\name{Maxime Amblard, Bruno Guillaume, Siyana Pavlova, Guy Perrier} 

\address{Université de Lorraine, CNRS, Inria, LORIA, F-54000 Nancy, France \\
         \{firstname.lastname\}@loria.fr\\}

\abstract{
This paper presents how the online tool \gm can be used to make queries and visualise data from existing semantically annotated corpora.
A dedicated syntax is available to construct simple to complex queries and execute them against a corpus.
Such queries give transverse views of the annotated data, these views can help for checking the consistency of annotations in one corpus or across several corpora.
\gm can then be seen as an error mining tool: when inconsistencies are detected, it helps finding the sentences which should be fixed.
Finally, \gm can also be used as a side tool to assist annotation tasks helping to find annotation examples in existing corpora to be compared to the data to be annotated.
 \\ \newline  }

\begin{document}

\maketitleabstract

\section{Semantic annotations as graphs}

There are a huge number of proposals in the literature to describe the formal representations of the semantics of natural language texts.
This diversity can be due to several factors; the main one being different linguistic theories used in the modeling.
We also observe differences in terms of levels of annotations or with a specific focus on some level.

Most of these representations use the notions of objects as \textit{entities} and \textit{events}. They describe semantic relations between these objects.
Of course, many propositions go further and propose other mechanisms to deal with temporal aspects or to describe the scope or the restriction linked to the logical interpretation of determiners as quantifiers; but we can consider that semantic relations between entities and/or events are a kind of minimal common denominator of the these proposals.

The mathematical notion of graphs is well-adapted to describe such kind of objects and we propose here to consider insofar as possible semantic annotations as graphs.
In our context, we consider labeled graphs, where nodes are decorated with flat features structures and edges are associated with specific labels.

In this paper, three semantic annotation frameworks are considered: Abstract Meaning Representation (AMR), Discourse Representation Structure (DRS), as they are used in the Parallel Meaning Bank (PMB), and QuantML.
The freely available annotated data for these three frameworks are now available in the \gm\footnote{\url{http://semantics.grew.fr}} tool. In the following, we briefly review these frameworks and illustrate how the tool facilitates annotation while making it more consistent.

Apart from tools specific to the different formalisms, we can cite~\newcite{cohen-etal-2021-repgraph} which also proposed a generic framework based on graph visualisation adapted to several semantic frameworks.
However, the tool is more focused on single graph visualisation and with manipulation features.
It does not propose complex queries with negative application patterns or the clustering feature we describe here for \gm.

\subsection{AMR}
The Abstract Meaning Representation (AMR)~\cite{banarescu2013} is a proposal whose focus is the predicate argument structure, using PropBank~\cite{palmer-etal-2005-proposition} as an inventory of semantic concepts.

\begin{figure}[h]
  \centering
\begin{small}
\begin{verbatim}
(r / resemble-01
  :ARG1 (y / you)
  :ARG2 (f / fox
    :poss (i / i))
  :time (k / know-02
    :ARG0 i
    :ARG1 f
    :ord (o / ordinal-entity :value 1)))
\end{verbatim}
\end{small}

  \includegraphics[scale=.35]{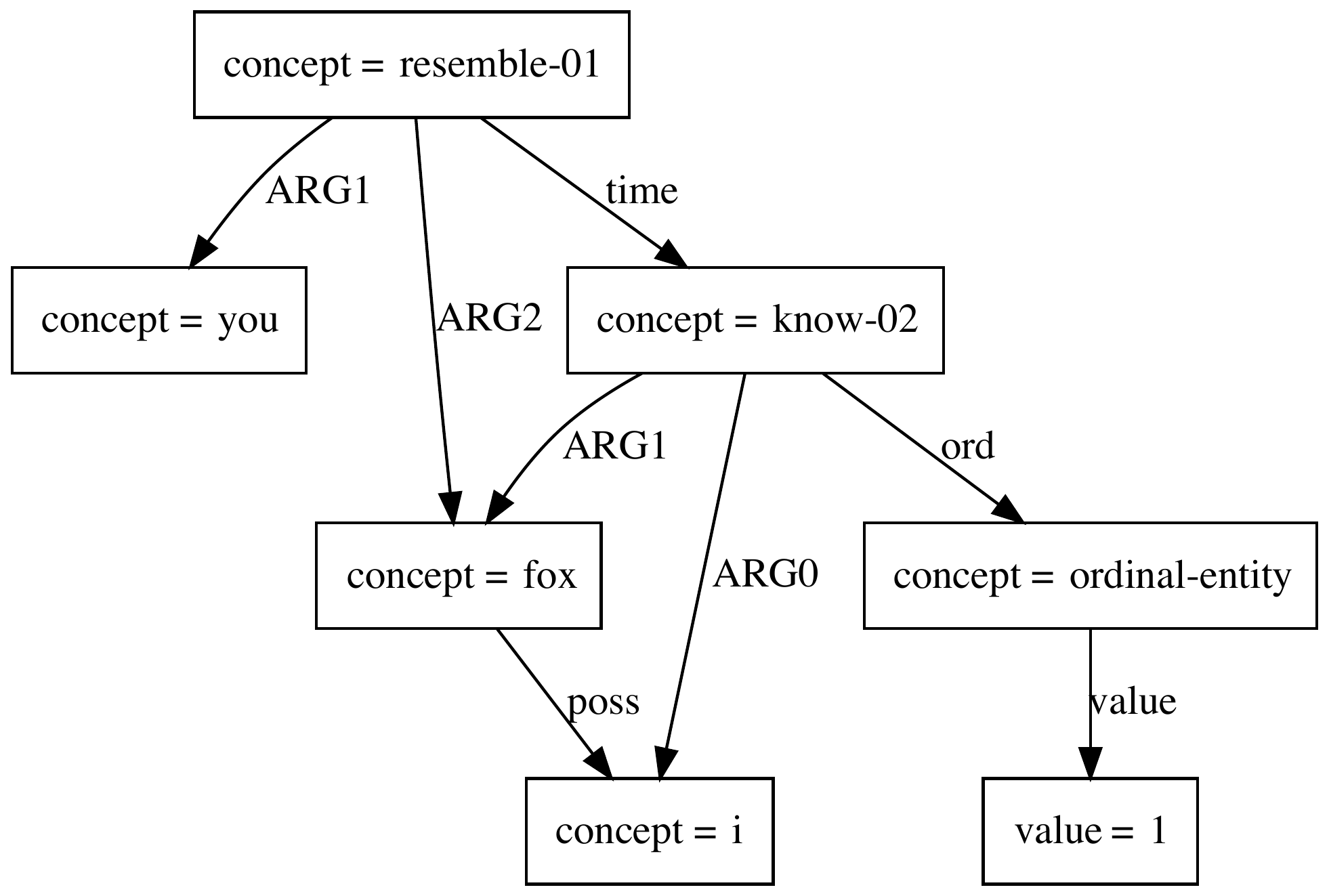}
  \caption{AMR annotation in Penman notation and its interpretation as graph of the sentence \texttt{[lpp\_1943.1161]}
  \emph{You are like my fox when I first knew him.}}
  \label{amr}
\end{figure}

As shown in \autoref{amr}, we interpret an AMR annotation as a graph in the following way:
\begin{itemize}
  \item each concept (like \emph{fox} or \emph{know-02} is a node with a feature name \texttt{concept};
  \item each value (like \emph{1}) is a node with a feature name \texttt{value};
  \item each semantic relation, with prefix ``:'' in Penman notation is an edge, typed with the relation name.
\end{itemize}

In \gm, two freely available AMR English corpora can be queried:
The Little Prince Corpus version 3.0\footnote{\url{https://amr.isi.edu/download/amr-bank-struct-v3.0.txt}} (1,562 sentences)
and BioAMR Corpus version 3.0\footnote{\url{https://amr.isi.edu/download/2018-01-25/amr-release-bio-v3.0.txt}} (6,952 sentences).

\subsection{DRS in the PMB}
There are several presentations of the DRS structures. 
In this paper we focus on the one used in the Parallel Meaning Bank (PMB)~\cite{PMB}\footnote{\url{http://pmb.let.rug.nl/}}, version 4.0.0, released in October 2021.
As in the AMR case, the predicate-argument structure is described with typed entities and typed semantic relations which can be converted into a graph representation. 
In addition, the box notation is used to describe the discourse relations and other constructions for which a notion of scope is needed (like quantifiers or negation).
The box notation requires a specific encoding into the graph structure.
Following the Bos' proposal~\cite{bos_semanticsarchive,pmb_sbn}, each box is drawn as a new node. Moreover, the embedding of a semantic node in a box is marked with a link which is drawn with a dotted line and labelled with the relation name \edge{in} in the figures.

\begin{figure}[h]
  \centering
\begin{small}
\begin{verbatim}
                  NEGATION -1
be.v.01           Theme 15 Co-Theme +1
prime_number.n.01
\end{verbatim}
\end{small}

  \includegraphics[scale=.35]{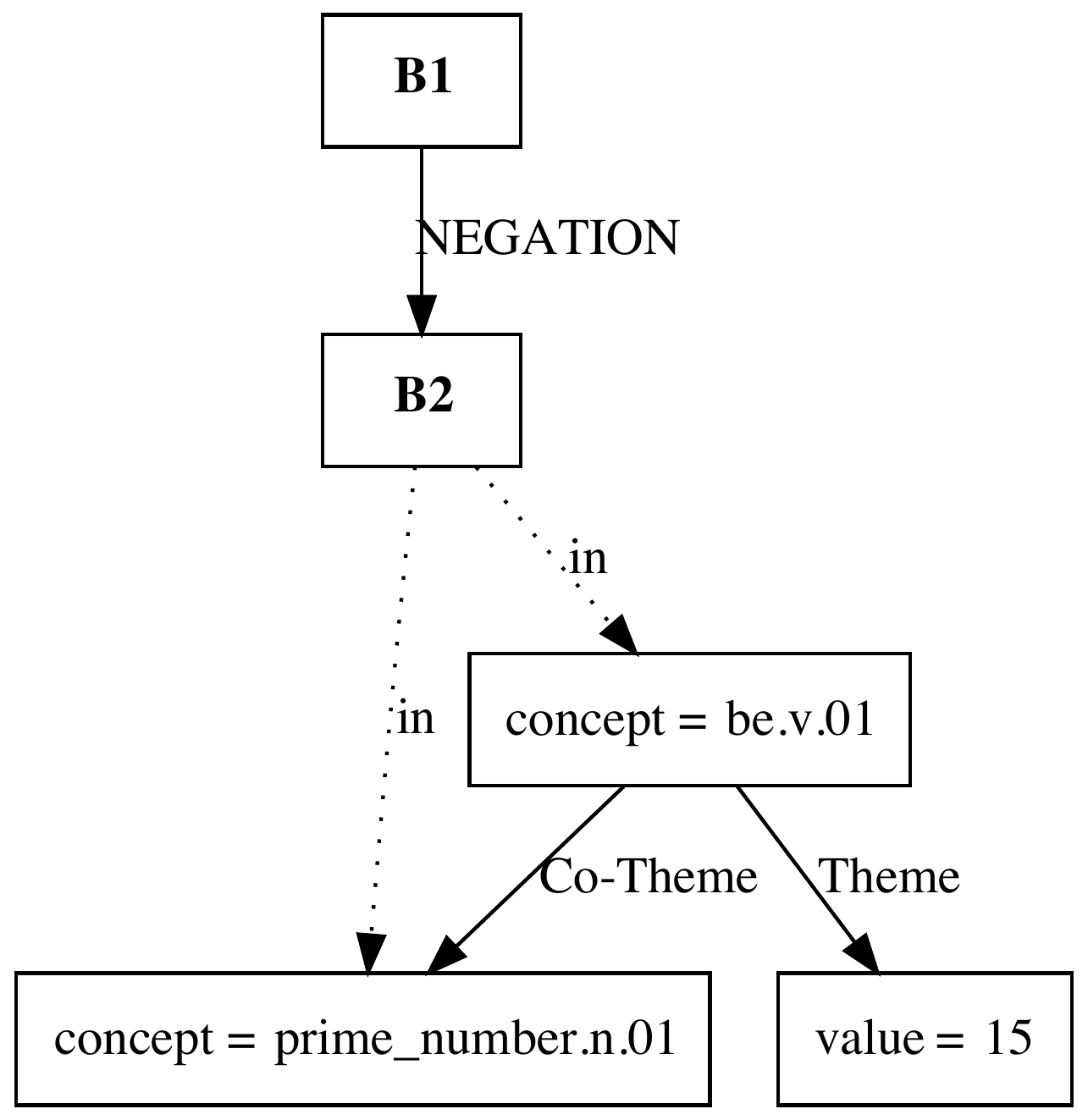}
  \caption{PMB annotation in SBN (Simplified Box Notation) and its interpretation as graph of the sentence \texttt{[p52/d2324]}
  \emph{Fifteen is not a prime number.}}
  \label{pmb}
\end{figure}

\autoref{pmb} shows an example of the a DRS annotation from the PMB and its representation as a graph.
In \gm, the gold data of the PMB is available (10,715 sentences in English, 2,844 in German, 1,686 in Italian and 1,467 in Dutch).

\subsection{QuantML}
QuantML~\cite{bunt-etal-2018-towards,bunt2020semantic} is another semantic annotation with a focus on quantification.
There is currently no annotated corpus in QuantML but a few annotations are proposed in the Guidelines part of the technical report~\cite{bunt2020semantic}. 
Again, the so-called \emph{Concrete syntax} of examples from the guidelines are converted into graphs.

\begin{figure}[h]
  \centering
  \includegraphics[scale=.35]{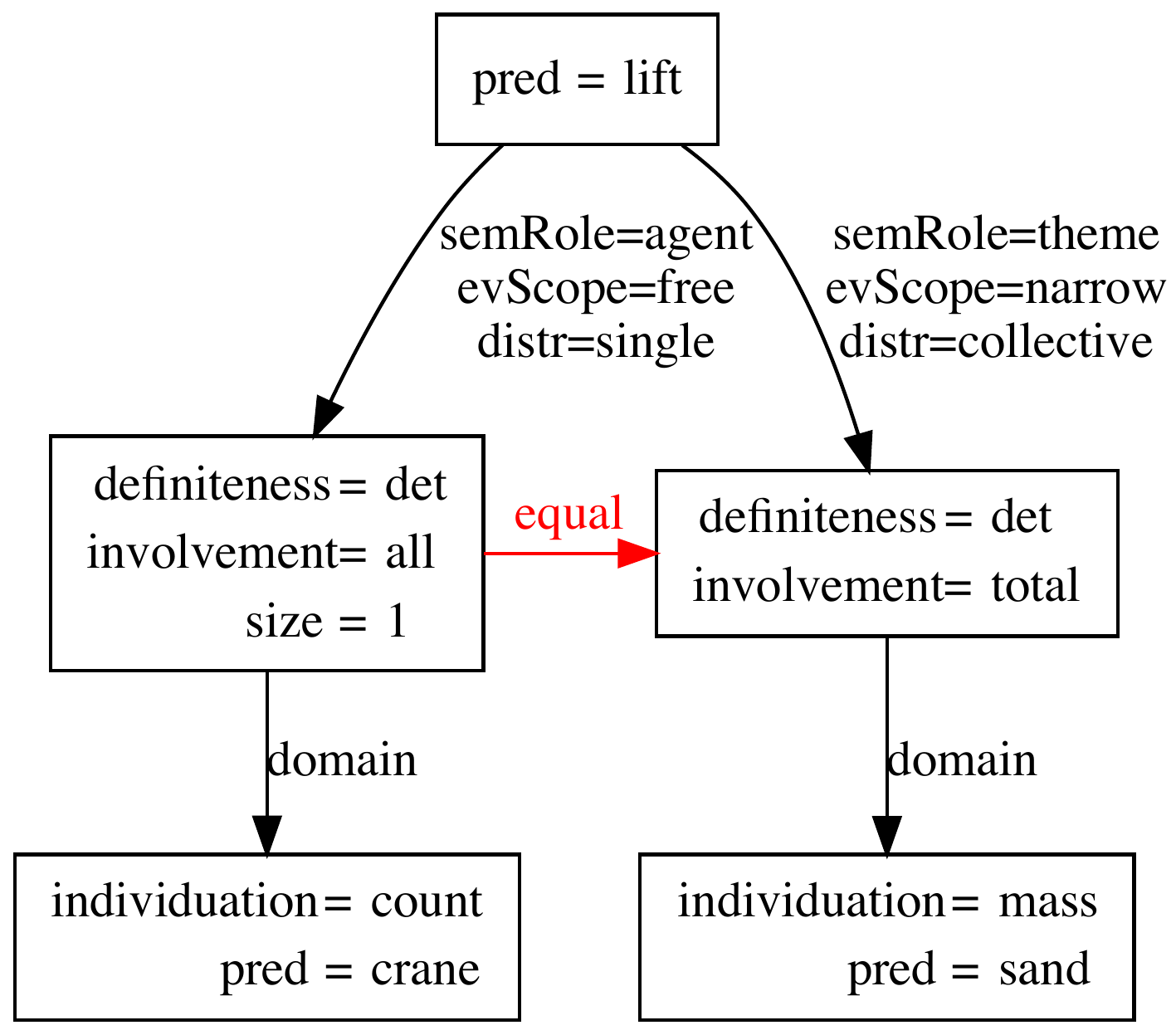}
  \caption{QuantML graph of the sentence \texttt{[A10]}
  \emph{The crane lifted all the sand}}
  \label{quantml}
\end{figure}

An example of the graph associated to the concrete level of a QuantML annotation is shown on \autoref{quantml}.
These graphs are richer: they use a skeleton with predicate/argument structure, but information about definiteness, distributivity or scope constraint is also given.
Features structures are used to describe different semantic aspects both on nodes and edges.
Scoping constraint between different arguments of the same predicate can be expressed (red edge \edge{equal} in the figure).

In~\newcite{amblard:hal-03298940}, we have participated in the ISA-17 shared task and proposed such annotations for 7 English sentences (with two alternative annotations for an ambiguous sentence).
In \gm, these annotations are available together with the 11 English examples\footnote{for five of them, we have detected annotation errors (see an example in \autoref{erro_mining} and we give both the original version and the fixed version.} of the guidelines in ~\newcite{bunt2020semantic}.

\section{The \gm tool}

In some previous works~\cite{bonfante:hal-01814386}, we have proposed to consider the representation of linguistic structures as graphs and to promote the well-studied computational model of Graph Rewriting to describe transformations of these structures.
In this framework, complex transformations can be encoded as a sequence of elementary and local transformations.
The local steps are graph rewriting rules, composed of a pattern (describing the part of the graph to be modified) and a sequence of commands (describing in which way the graph is changed).
The \grew~\cite{guillaume:hal-03177701} tool is an implementation of this framework.

A mechanism of graph matching is used in \grew to detect when a rule can be applied to a graph.
But we have observed that this mechanism can be used on its own as a way to query a graph or a collection of graphs.
This querying aspect give birth to a new tool, called \gm, a web-based interface to express queries on annotated corpora and to visualise the occurrences returned by the query.

\begin{figure*}
  \centering
  \includegraphics[scale=.3]{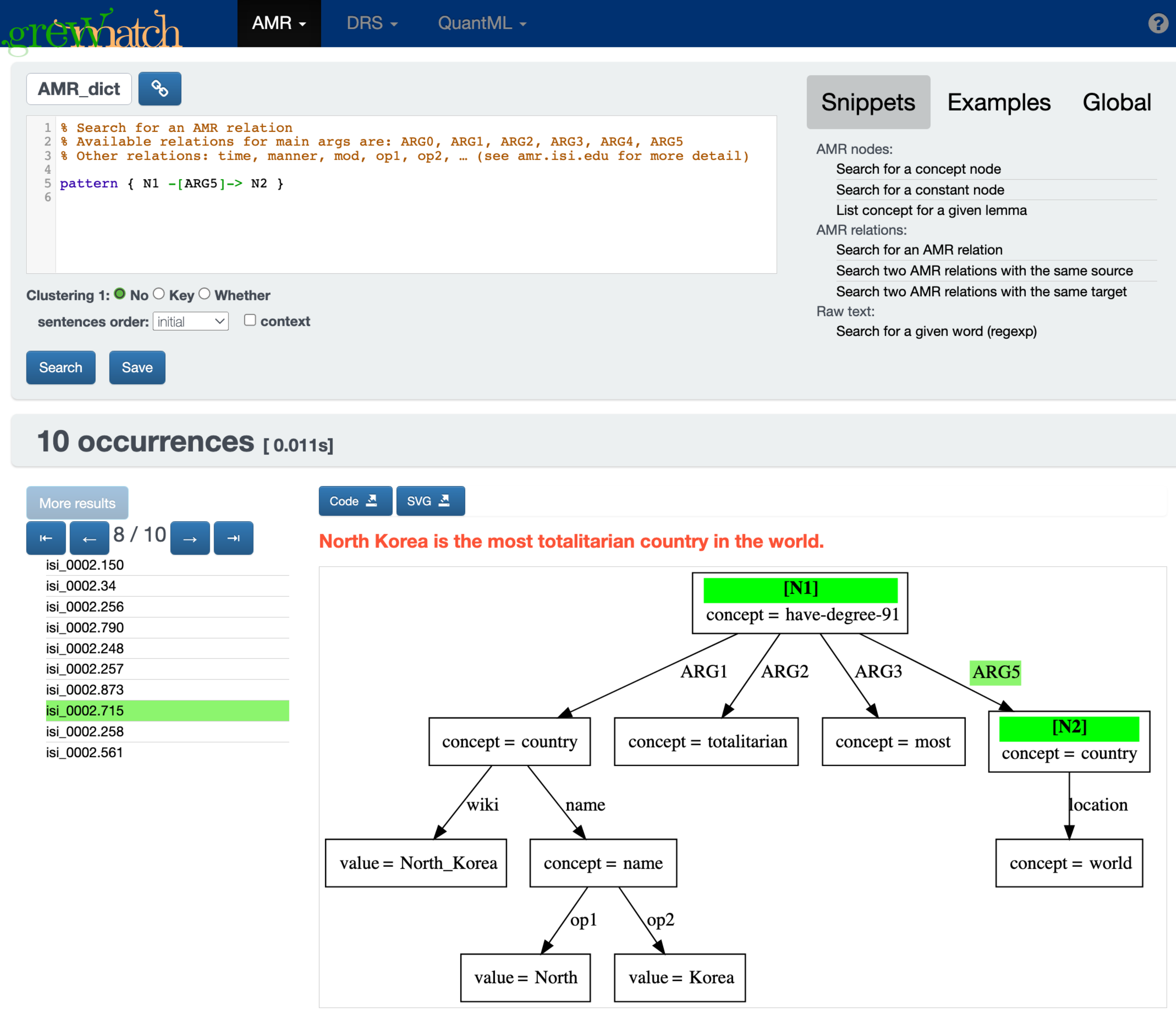}
  \caption{The \gm tool}
  \label{gm}
\end{figure*}

A screenshot of the tool applied to AMR example is shown in \autoref{gm}.
The visualisation of semantic structures uses the Graphviz tool\footnote{\url{https://graphviz.org/}}.

Theoretically, graph matching is an NP-complete problem but in the present context, matching is done on a set of small graphs (one graph per sentence) and then the complexity is not an issue and most of the graph requests can be executed very quickly.
We have made a few experiments on larger graphs (around 20,000 nodes) and then the complexity strongly relies of the shape of the pattern.
If a pattern has a tree structure, the matching is easing and linear in the size of the graph; for general patterns, there is no generic efficient algorithm and some heuristics will be needed.

\subsection{The query language}

We briefly describe here the main aspects of queries in \gm, we let the reader go to the \grew documentation pages for more details.
The main part of a query is introduced by the keyword \texttt{pattern} which describes the set of nodes and edges that should be matched in the host graph.
For nodes and edges, several constraints can be expressed. More general constraints can also be expressed, for example the fact different nodes share the same feature value. 
Given a corpus and a basic request (introduced by the keyword \texttt{pattern}), we can further refine the query by adding negative application patterns (introduced by the keyword \texttt{without}). 
Each negative application pattern is a constraint that filters out the occurrences returned by the basic pattern according to additional constraints.

As an example, the following pattern with one basic pattern (first line) and two negative application patterns (last two lines).
It shows the syntax to express: find all the concept nodes in the corpus where the concept is \texttt{say-01} but such that there are no outgoing edges labelled \edge{ARG0} from this node (note that the identifier \texttt{N} is used to refer to the same node) and such that there are no incoming edges labelled \edge{ARG0-of} on this node.
An example of a graph returned by this pattern is given in \autoref{sec:ling} below.

\begin{small}
\begin{verbatim}
  pattern { N [concept = "say-01"] }
  without { N -[ARG0]-> A0 }
  without { A0 -[ARG0-of]-> N }
\end{verbatim}
\end{small}

Another feature which has been proved useful in \gm is the ability to cluster the results given by a complex request.
The user can chose a clustering key (like the feature of one of the nodes of the basic pattern), the set of occurrences is clustered according to value of this feature (see example in the next section).

The clustering can also be done following a sub-pattern: considering a pattern $P$ and a sub-pattern $P'$, all occurrences of $P$ are clustered in two subsets $P_{yes}$ and $P_{no}$ depending on whether $P'$ is also satisfied by the considered occurrence.
For instance, we can observe how coordination is annotated with the pattern $P$: \verb+pattern { N [concept = "and"] }+ and the sub-pattern \verb+N -[op1]-> X+, to see if the concept \texttt{and} appears with or without an \edge{op1} outgoing edge. On The Little Prince, there are 215 occurrences in $P_{yes}$ and 127 in $P_{no}$.
With the same $P$ and the sub-pattern \verb+N -[op2]-> X+, the occurrences are 240 in $P_{yes}$ and 102 in $P_{no}$.
This shows that ``unary'' coordination (sentence beginning with the word \emph{and} are not consistently annotated: the unique conjoint is sometimes annotated \edge{op1} and sometimes \edge{op2}.

\section{Linguistic observations on semantic annotations}
\label{sec:ling}

We list here a few examples of requests which can be used to make observations on the annotated corpora.

\paragraph{Concepts linked to a given verb.}
With the following request and a clustering on \texttt{N.concept}
\begin{small}
\begin{verbatim}
  pattern { N [concept = re"make-.*"]; }
\end{verbatim}
\end{small}

we obtained the distribution of the usage of the concepts.
On The Little Prince Corpus, the concepts returned are \texttt{make-02} (18), \texttt{make-01} (17), \texttt{make-05} (1), \texttt{make-06} (1) and  \texttt{make-up-07} (1).

\paragraph{Realisation of an argument of a predicate.}
In The Little Prince Corpus, the most frequent predicate is \texttt{say-01} (234 occurrences).
According to PropBank, this predicate has 4 core arguments: \edge{ARG0} (\textit{Sayer}), \edge{ARG1} (\textit{Utterance}), \edge{ARG2} (\textit{Hearer}) and \edge{ARG3} (\textit{Attributive}).
With a few requests on \gm, we can observe how often the different arguments are realised of not. 
For \edge{ARG0}, the following request gives the 6 occurrences of the predicate without the \textit{Sayer}.
Note that we have to take care both of the \edge{ARG0} outgoing edges (line 2) but also to the \edge{ARG0-of} incoming edges (line 3) (without the last line, 9 occurrences would be wrongly reported).

\begin{small}
\begin{verbatim}
  pattern { N [concept = "say-01"] }
  without { N -[ARG0]-> A0 }
  without { A0 -[ARG0-of]-> N }
\end{verbatim}
\end{small}

\begin{figure}[h]
  \includegraphics[scale=.35]{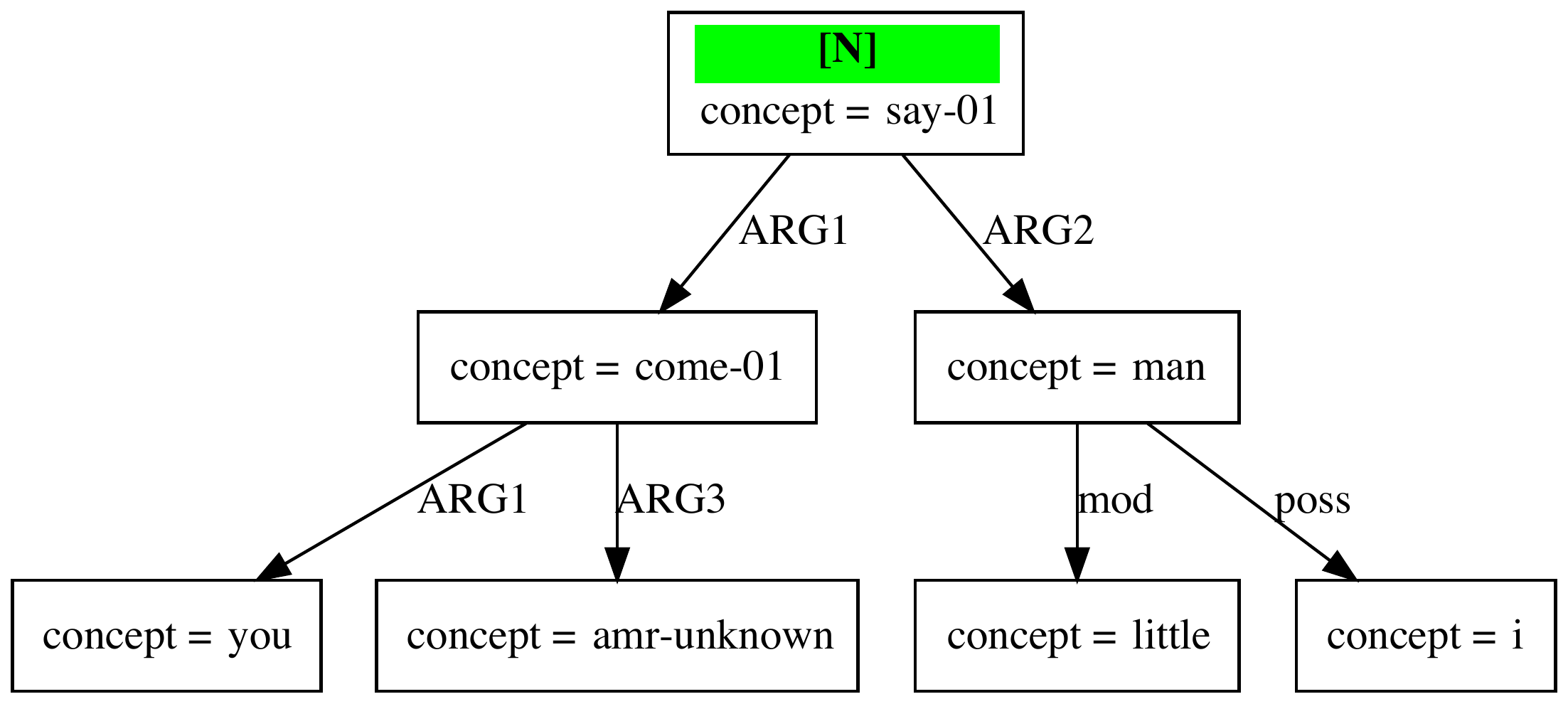}
  \caption{AMR annotation of the sentence \emph{"My little man, where do you come from?}.}
  \label{say_no_arg0}
\end{figure}

An example of one of the six occurrences is shown in \autoref{say_no_arg0}.

\begin{figure}[h]
  \centering
  \includegraphics[scale=.5]{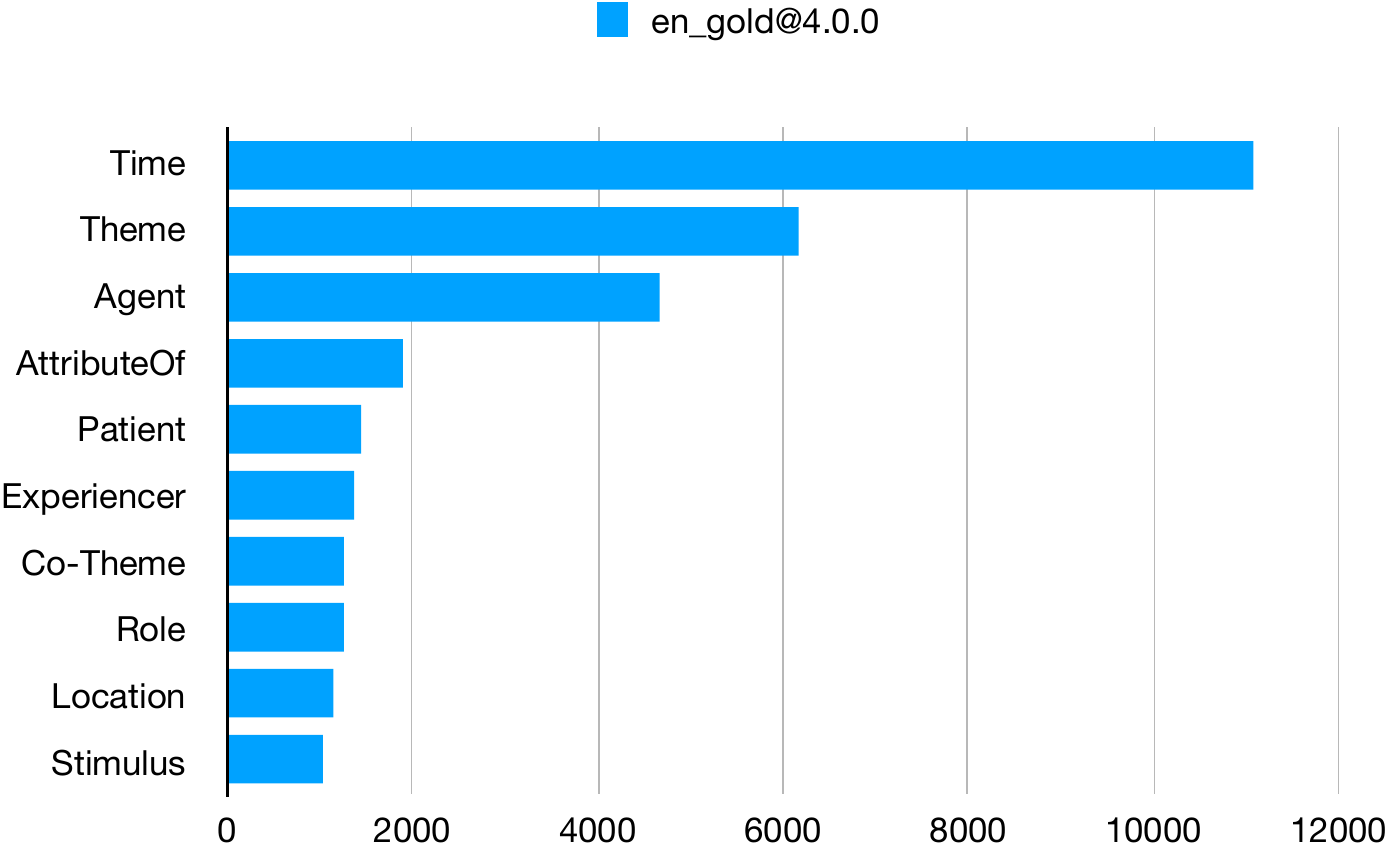}
  \caption{distribution of the ten most frequent semantic relations between two ``concept'' nodes}
  \label{sem_stat}
\end{figure}

\paragraph{Observation of distributions in the data.}
Graph querying is also available through scripts which produces statistics about the number of occurrences in corpora.
With the following pattern, and a clustering following the label of the edge named \texttt{e}, we can observe the distribution of relations between two ``concept'' nodes (see \autoref{sem_stat}).

\begin{verbatim}
  pattern { 
    M [concept]; N [concept]; 
    e: M -> N 
  }
\end{verbatim}

With the encoding of boxes we have described above, it is possible to request for specific configurations of boxes.
The next pattern corresponds to two nested negations.

\begin{small}
\begin{verbatim}
  pattern { 
    B1 -[NEGATION]-> B2;
    B2 -[NEGATION]-> B3
  }
\end{verbatim}
\end{small}

\begin{figure}[h]
  \centering
  \includegraphics[scale=.35]{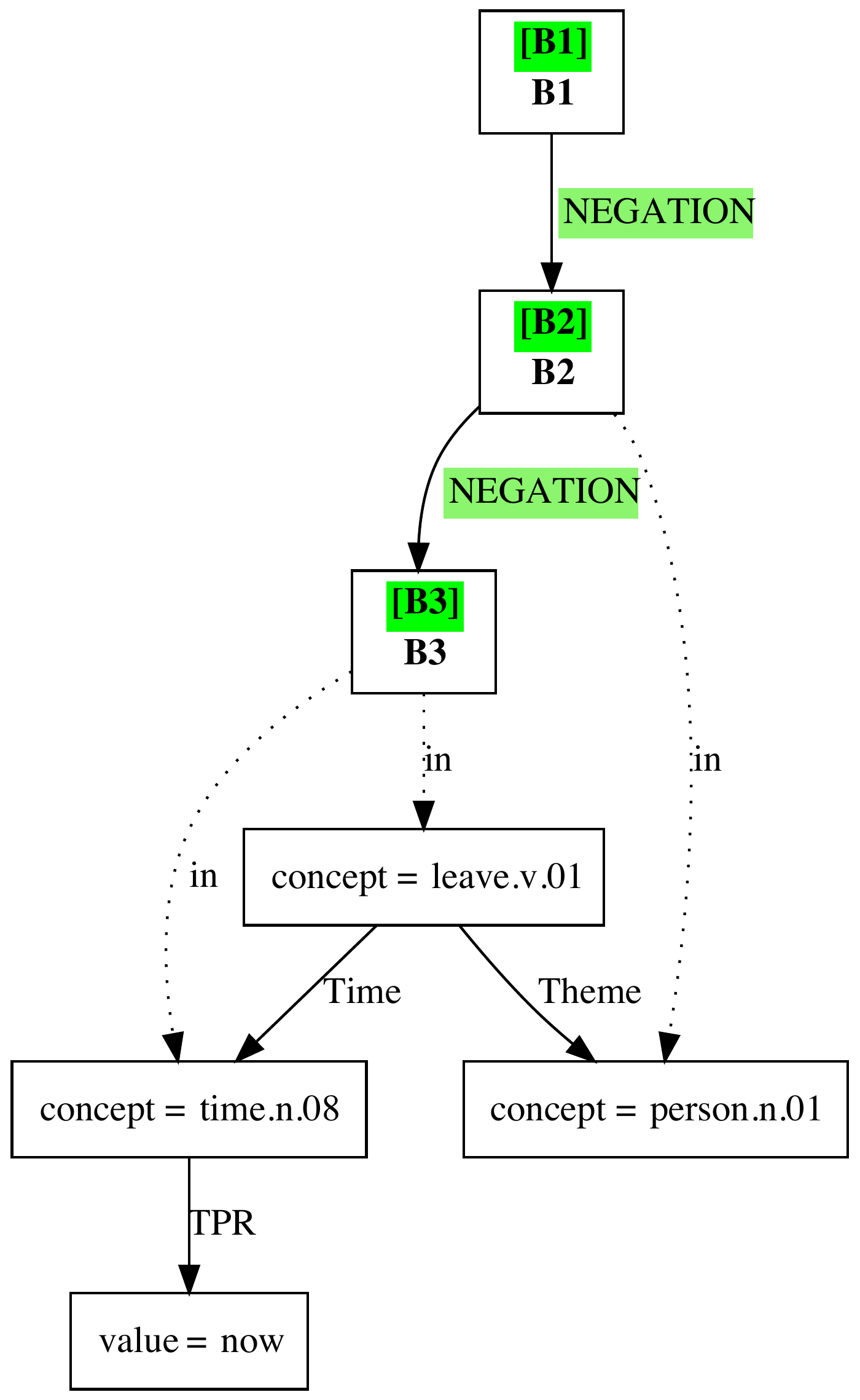}
  \caption{Double negation encoding for the sentence \texttt{[p18/d1454]} \emph{Everybody left}.}
  \label{double_neg}
\end{figure}

With this query, in \autoref{double_neg}, we can observe a perfect illustration of the encoding of universal quantification through a double negation.\footnote{In PMB, TPR stands for \emph{temporal precedence.}}

\paragraph{Global graph structure.}
More general queries about the graph structure allow for instance to check for cyclic structures.

\begin{small}
\begin{verbatim}
  global { is_cyclic }
\end{verbatim}
\end{small}

The AMR guidelines say ``Approximately 0.3\% of AMRs are legitimately cyclic''%
\footnote{\url{https://github.com/amrisi/amr-guidelines/blob/master/amr.md}}.
But, with the query above, we can report the ratio of cyclic structures in AMR corpora: more than 3\% in The Little Prince Corpus and almost 6.9\% in the BioAMR corpus.

On the gold data of the PMB, cyclic structure are rare: there are 34 cyclic structures in English (among 10,715 sentences) and 1 in German (in 2,844 sentences).
There are no examples in Italian or Dutch data.

\section{Error mining}
\label{erro_mining}

\gm can be used to detect inconsistencies in the annotations.
A query is designed to express a constraint which should be respected by all the annotated structures.
Such a query is supposed to return an empty set of occurrences.
If it is not the case, we can observe the exceptions given.
These can be annotation errors or if the annotation is legitimate, the query should be refined to take into account these cases.
It also helps to find missing information in the guidelines where some cases are not recorded.

We give below a few examples of such usage of \gm for inconsistencies detection.

In AMR structures, according to the guidelines, each named entity, is annotated with a node whose concept expresses the kind of entity (\emph{Person}, \emph{City}\ldots) and with two outgoing edges labeled \edge{name} and \edge{wiki}.
With the following pattern, we can search for nodes with an outgoing edge \edge{name} and without an \edge{wiki} edge, and spot inconsistent annotations.

\begin{verbatim}
  pattern { M -[name]-> N }
  without { M -[wiki]-> * }
\end{verbatim}

This pattern returns one occurrence in the data from the AMR Annotation Dictionary\footnote{\url{https://www.isi.edu/~ulf/amr/lib/amr-dict.html}, consulted on 2022/03/31} where the city name \emph{New Orleans} is not associated with its wikipedia page.
We can also report that the BioAMR Corpus is not consistently annotated in this respect: 95\% of \edge{name} edges appear without a \edge{wiki} edge.

\begin{figure}[h]
  \centering
  \includegraphics[scale=.35]{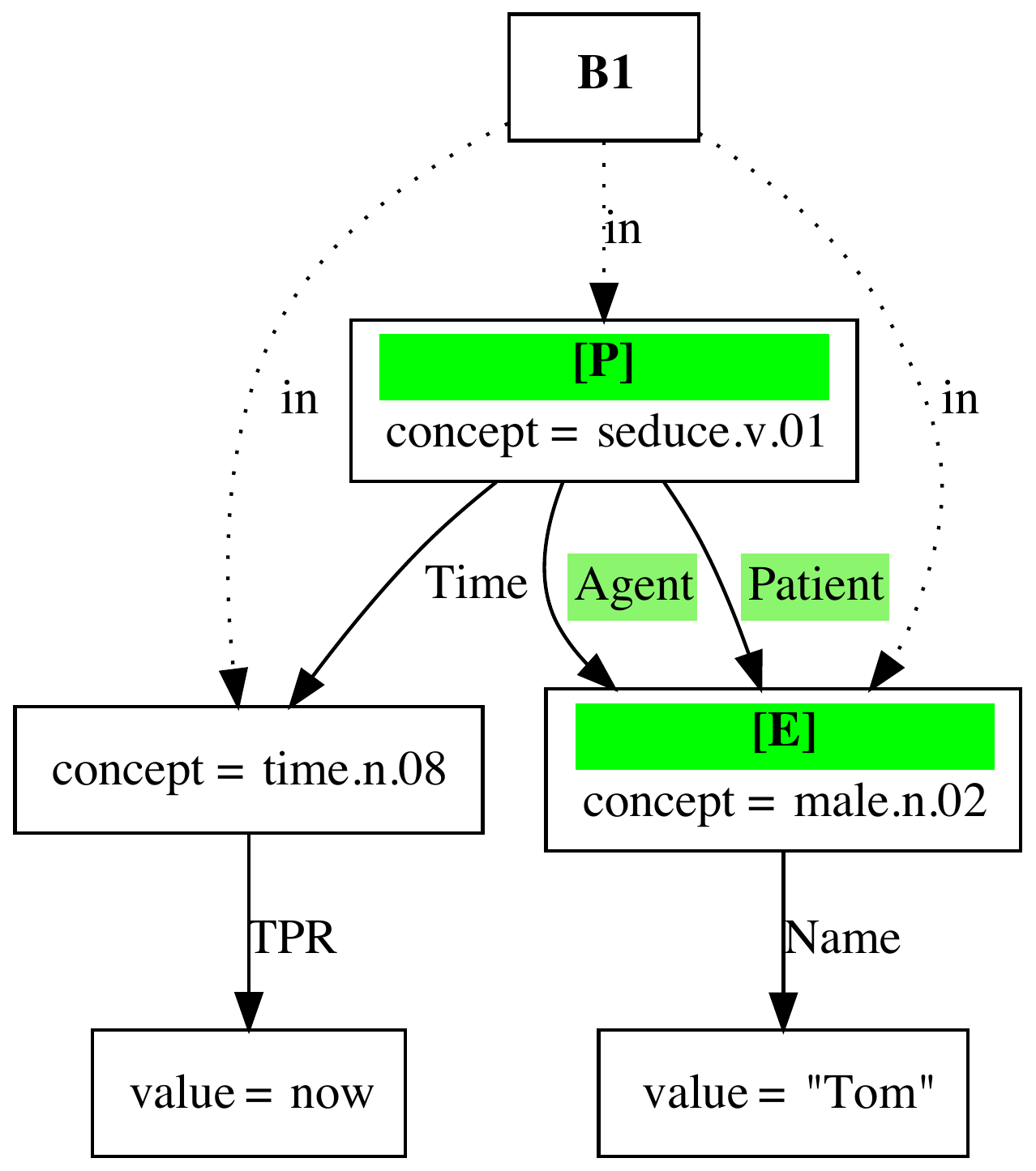}
  \caption{PMB annotation of the sentence \texttt{[p62/d1397]} \emph{He was seduced by Tom}.}
  \label{p62_d1397}
\end{figure}

On the PMB, we can use the following pattern to observe structures where the same entity (node E) is both the \edge{Agent} and the \edge{Patient} of the same predicate P.
\begin{small}
\begin{verbatim}
  pattern {
    P -[Agent]-> E;
    P -[Patient]-> E;
  }
\end{verbatim}
\end{small}
On the English gold data (10,715 sentences), 20 occurrences are returned.
In 15 cases, the pattern is legitimate (sentences with \emph{himself, herself\ldots}) but the 5 remaining cases are annotation errors: for instance, \texttt{[p60/d0784]}\emph{Betty killed her mother.} or \texttt{[p62/d1397]}\emph{He was seduced by Tom.} (see \autoref{p62_d1397} for this last sentence).

\begin{figure*}
  \centering
  \includegraphics[scale=.35]{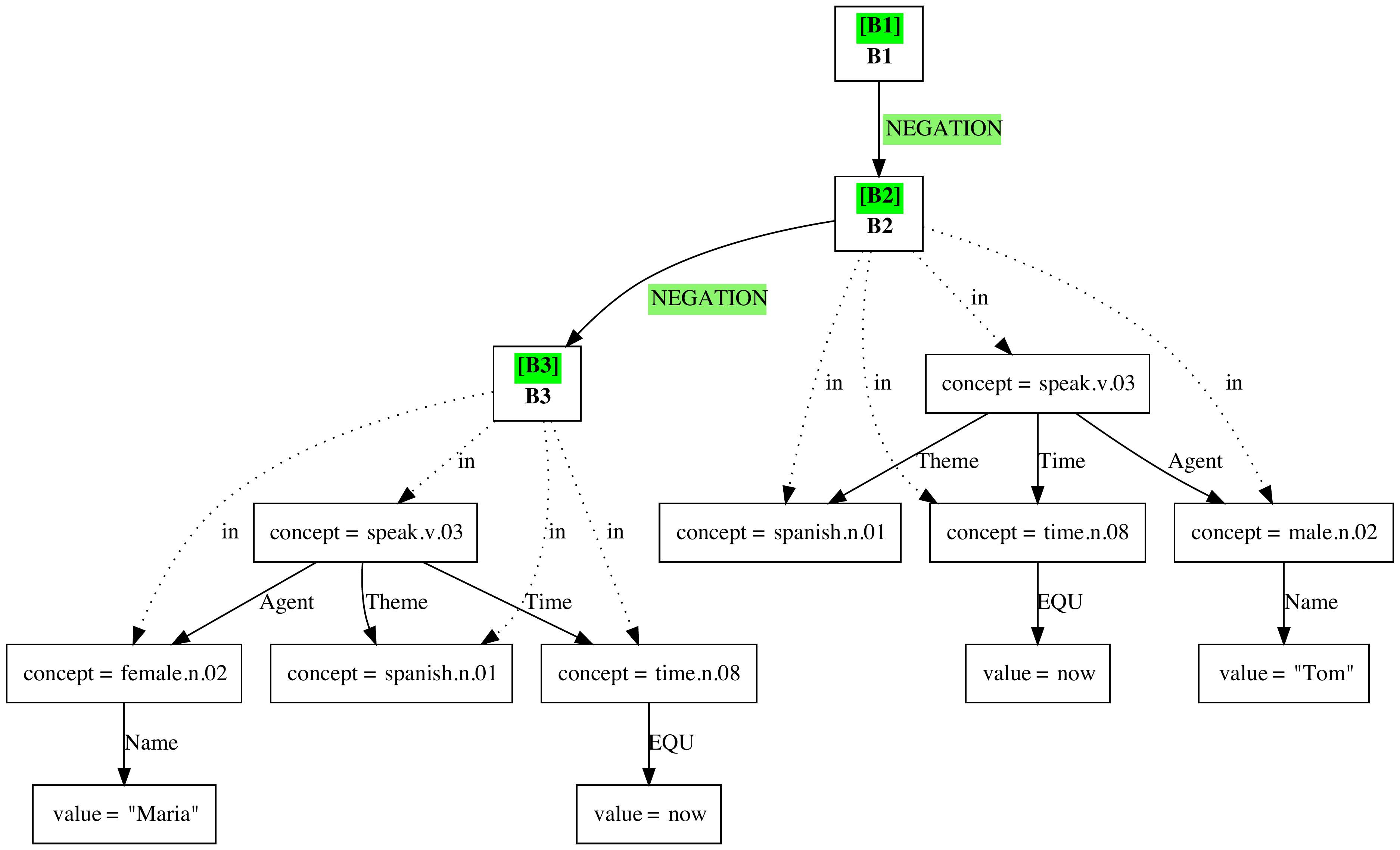}
  \caption{SBN annotation in the PMB of the German sentence \texttt{[p38/d2263]} \emph{Weder Tom noch Maria sprechen Spanisch} (\emph {`Neither Tom nor Mary speaks Spanish.'}).}
  \label{bug_pmb}
\end{figure*}

With the pattern already given in \autoref{sec:ling} for double negation, we retrieve also the example given in \autoref{bug_pmb} where the structure is not the one expected.
The two internal boxes should be at the same level and not embedded as in the figure.
In fact, in the clause notation of the PMB (the original notation from which the SBN notation is extracted), the sentence is correctly annotated.
We have indeed found a bug in the conversion process for the SBN notation which has been reported to the PMB maintainers.

\begin{figure}[!htbp]
  \centering
  \includegraphics[scale=.35]{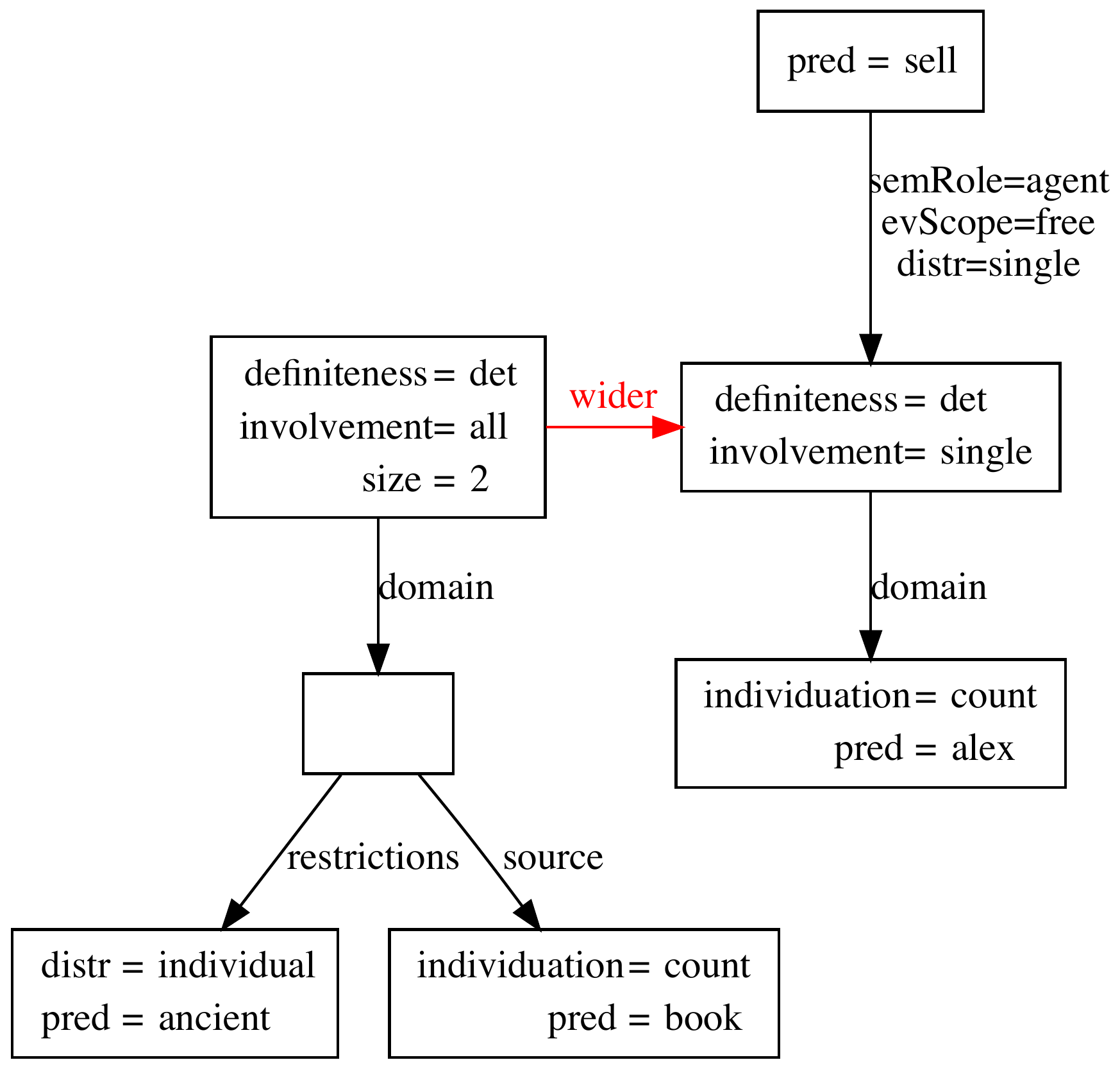}

  \vspace{10mm}
  \includegraphics[scale=.35]{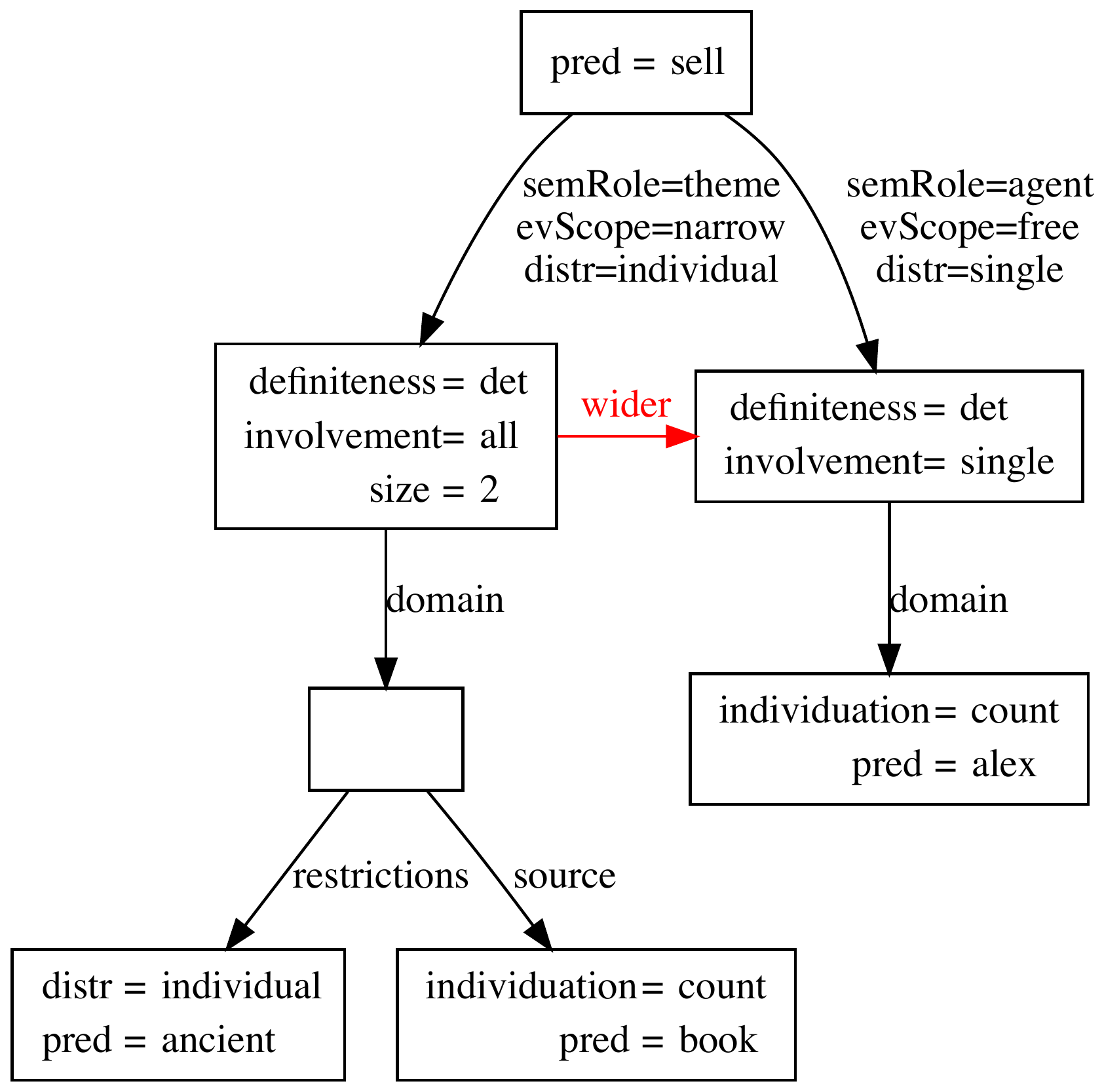}
  \caption{QuantML annotation of the sentence \texttt{[A7]} \emph{Alex sold the two ancient books} in the TiCC report (top) and the corrected annotation (bottom).}
  \label{quantml-bug}
\end{figure}

On QuantML, the number of available annotated sentences is really tiny: 11 sentences in the TiCC report and 7 sentences in \newcite{amblard:hal-03298940}.
Hence, sentences can be checked one by one without using queries;
nevertheless, having a graph visualisation of this annotation was useful.
When working on the ISA-17 shared task, we started producing the graphs for the examples in the guidelines and we discovered some inconsistencies.
In \autoref{quantml-bug}, we present the wrong graph of one example and the corrected version of the same annotation.

\section{Conclusion}

Semantic structures are often complex and represent several different levels of information in the same structure.
It is then very useful to provide graphical visualisation in order to assist the humans who have to work with these data, either as annotators or as users.
With the \gm tool, we propose to use the mathematical model of graphs as a common way to represent the semantically annotated data in various frameworks.
Doing this, we have the possibility to visualise the annotations but also to use the graph query languages provided to make request on corpora of annotated sentences.

Querying graphs with \gm has been useful to make linguistic observations on the data or to check the consistency of the data and the conformity with the guidelines.
When inconsistencies are reported, it helps finding how the data or the guidelines (or some other tool) should be improved.
\gm can also be used as a side-tool when doing annotation, which helps finding similar examples in the already annotated data and thus helps annotators to take consistent decisions for similar constructions.
We would like to recommend to use the methodology presented in this paper, based on graph visualisation and graph querying  as a non regression evaluation tool for any framework.

In future work, we plan to consider other semantic annotations frameworks like UCCA~\cite{abend-rappoport-2013-universal} or DMRS~\cite{copestake-2009-invited} for instance for which a graph based visualisation and querying would probably be useful as well.

\section{Acknowledgements}
This work was partially supported by the ANR fund (ANR-20-THIA-0010-01)
We would like to thanks Clara Serruau for her preliminary work on the subject and the reviewers for thier comments and suggestions.

\section{Bibliographical References}\label{reference}

\bibliographystyle{lrec2022-bib}
\bibliography{2022.isa18-1.13}

\end{document}